\documentclass[lettersize,journal]{IEEEtran}
\usepackage{amsmath,amsfonts}
\usepackage{algorithmic}
\usepackage{array}
\usepackage[caption=false,font=normalsize,labelfont=sf,textfont=sf]{subfig}
\usepackage{textcomp}
\usepackage{stfloats}
\usepackage{url}
\usepackage{pifont}
\usepackage[colorlinks]{hyperref}  
\usepackage[dvipsnames,table,xcdraw]{xcolor} 
\usepackage{multirow}
\usepackage{multicol}
\usepackage{verbatim}
\usepackage{cite}
\usepackage{graphicx}
\usepackage{caption}

\def\BibTeX{{\rm B\kern-.05em{\sc i\kern-.025em b}\kern-.08em
    T\kern-.1667em\lower.7ex\hbox{E}\kern-.125emX}}
\usepackage{balance}

\let\oldtwocolumn\twocolumn
\renewcommand\twocolumn[1][]{%
    \oldtwocolumn[{#1}{
    \begin{center}
           \includegraphics[width=\textwidth]{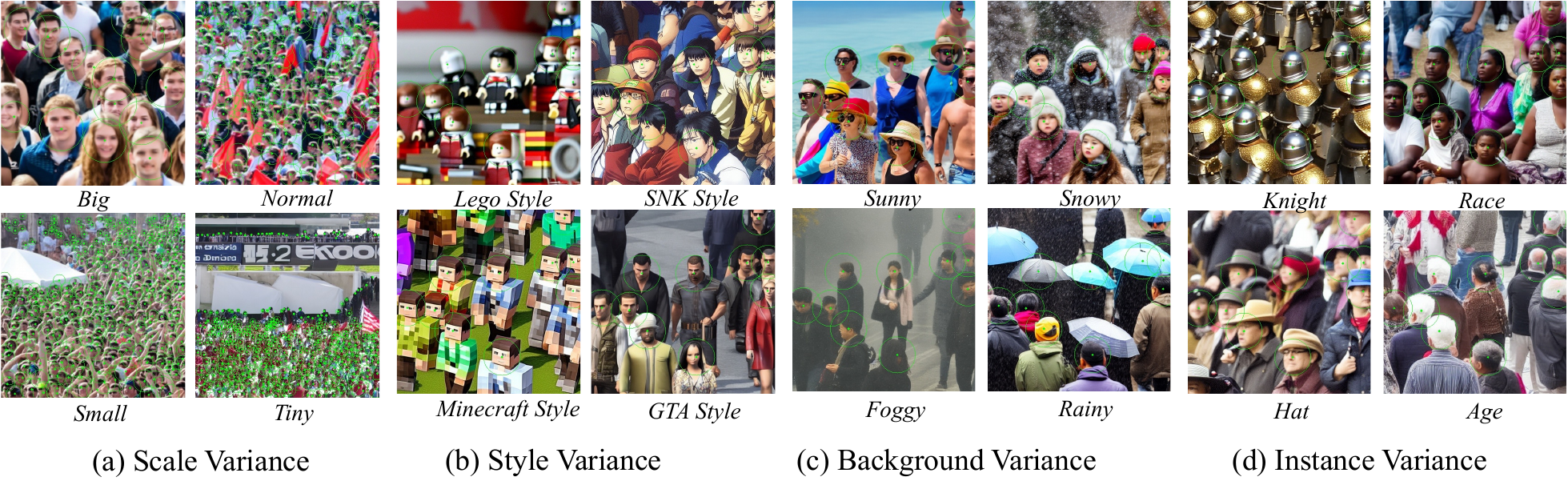}
           \captionof{figure}{\textbf{DenseControl is able to control the instance position and scale} in generating crowd images under 
           arbitrary scenarios, including different (a) scales, (b) styles, (c) backgrounds, and (d) instances. Best viewed zoomed in and in color.}
           \label{fig:teaser}
        \end{center}
    }]
}

\begin{document}
\title{\includegraphics[scale=0.1]{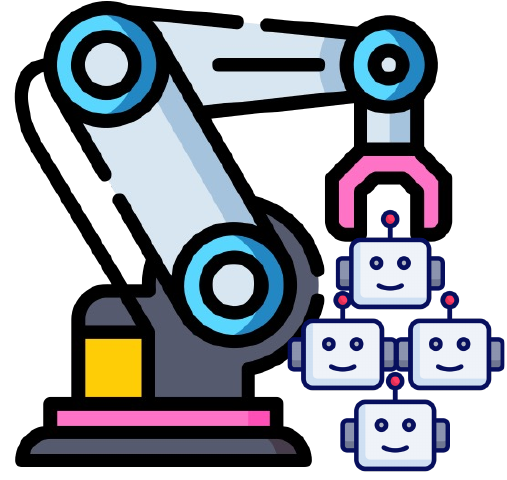} DenseControl: Instance-Level Controllable \\Synthesis of Dense Crowd Image }

\author{Juncheng Wang, Lei Shang, Wang Lu, Baigui Sun, Shujun Wang
\thanks{Corresponding to: Shujun Wang.\\
    Juncheng Wang is with the department of Biomedical Engineering, the Hong Kong Polytechnic University, Hong Kong (email: wjc2830@gmail.com);\\
    Lei Shang, and Baigui Sun are with Tongyi lab, Alibaba Group, Hangzhou, China;\\
    Wang Lu is with Tsinghua University, Beijing, China;\\
    Shujun Wang is with the Department of Biomedical Engineering, the Hong Kong Polytechnic University, Hong Kong SAR, China.}}

\markboth{Journal of \LaTeX\ Class Files,~Vol.~18, No.~9, September~2020}%
{How to Use the IEEEtran \LaTeX \ Templates}

\makeatletter
\newcommand\whline{\noalign{\ifnum0=`}\fi\hrule \@height 1.25pt \futurelet
	\reserved@a\@xhline}
\makeatother
\newcolumntype{C}{>{\collectcell\uline}c<{\endcollectcell}}

\maketitle

\begin{abstract}
In this paper, we introduce DenseControl, a novel pipeline for generating dense crowd images.
Specifically, DenseControl meticulously positions and sizes each generated instance to align precisely with the predefined coordinates and scales. Based on this, we further allow for control over the background, style, and attributes of instances.
The motivation behind DenseControl stems from the observation of two main challenges in synthesizing crowd images: controlling signal embedding and maintaining topological integrity when imparting instance scale guidance. 
To address these, we first introduce the Isolated Object Embedding (IOE) map, a novel representation that facilitates spatial location control while mitigating the difficulties associated with learning projections for model. 
Secondly, we propose an Implicit Scale Embedding (ISE) strategy that seamlessly integrates with the IOE map to encode precise scale information.
To further enhance the efficacy of combining ISE with the IOE map, we incorporate a Position Shortcut mechanism that enhances cross-attention to alleviate projection challenges. 
We evaluate DenseControl through two lenses: synthesis quality and applicability in latent applications. 
Experiments across different control conditions demonstrate DenseControl achieves state-of-the-art results in dense crowd image synthesis.  
Furthermore, we showcase applications in augmenting crowd analysis under data scarcity, transfer learning, and weather generalization scenes, to highlight the practical utility of DenseControl. 
The codebase will be released. 
\end{abstract}


\section{Introduction}\label{sec:introduction}

Crowd analysis~\cite{LSC-CNN, Tiny-Faces,RAZNet,LiangFIDTM, TopoCount, IIM, ScopedTeacher, CrowdHat, Hantao} is an essential field in the realm of computer vision, significantly influencing urban planning and public safety. 
One of the key obstacles in crowd analysis is the dense clustering of individuals within a space. Addressing this issue typically requires a collection of varied and extensive training datasets that can contain dense crowds and adequately cover various scenarios and environmental conditions. 
However, collecting and annotating high-quality, extensive crowd datasets are expensive and logistically complex, especially under challenging conditions like adverse weather or rare events.
Moreover, ethical and privacy concerns add another layer of difficulty in gathering sensitive data. Reliance on publicly available datasets can introduce biases and limitations, as these datasets may not adequately reflect the real-world diversity of crowd scenes. 
Consequently, overcoming the challenges of obtaining extensive, cost-effective, yet high-quality data is crucial for the advancement of automated crowd analysis.

\begin{figure}[t]
    \centering
    \includegraphics[width=0.5\textwidth]{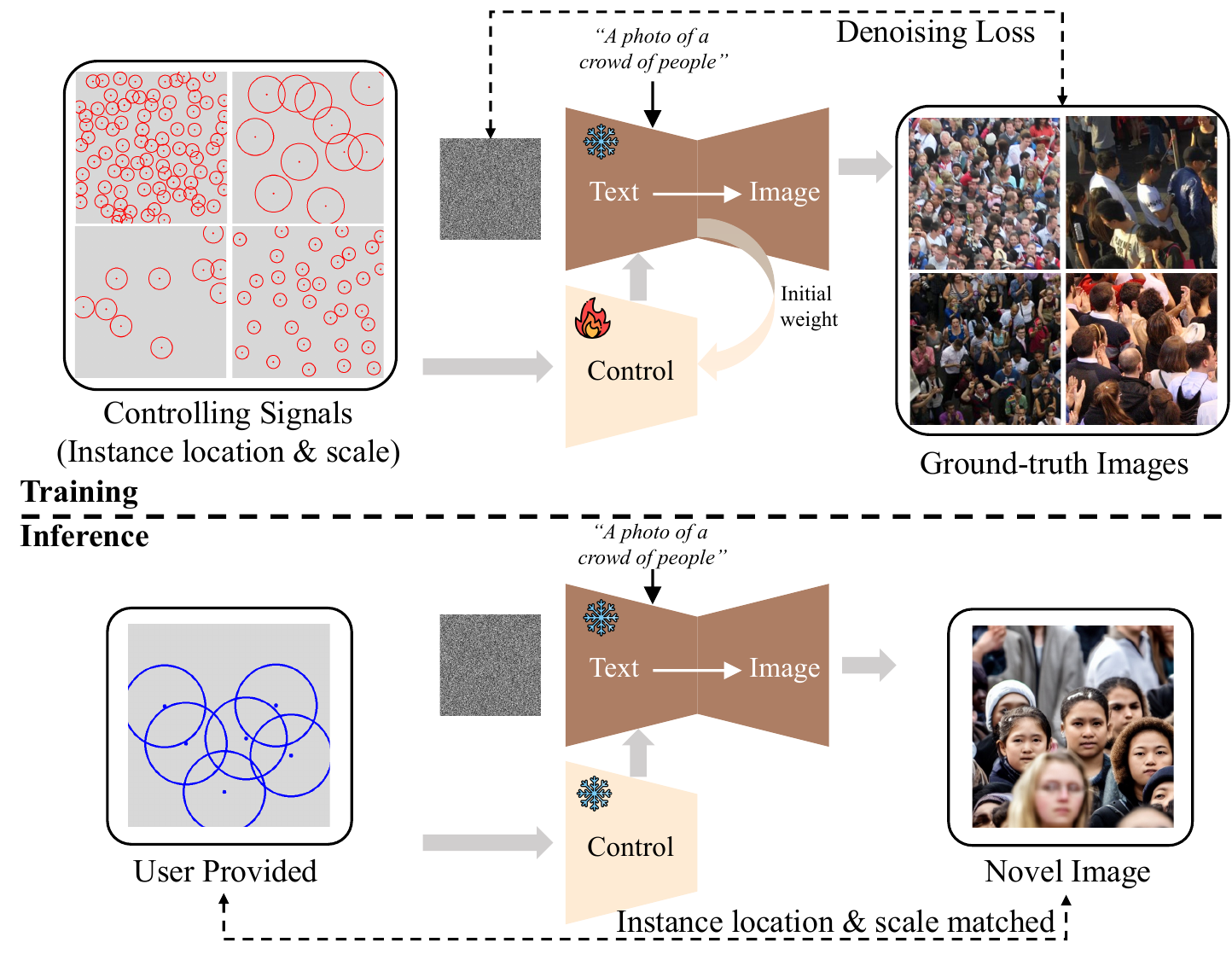}
    \caption{The DenseControl Pipeline Overview. During the training phase, DenseControl is fed with pairs of crowd images and corresponding control signals. The model employs the conventional diffusion denoising loss to complete its training. Upon completion, users can specify arbitrary combinations of locations and scales to generate new, customized images.}
    \label{fig:framework}
\end{figure}
To this end, generating synthetic crowd images with user pre-defined annotation could be the solution.
Previous work like GCC~\cite{GCC} proposes to utilize electronic game engine to create simulated crowd datasets. However, the realism of game engine-generated data remains questionable, as it relies on predefined instances and scenes determined by the game's developers, thus failing to satisfy our specific requirements.
In the search for more adaptable solutions, the potential of generative models, such as Stable Diffusion \cite{SDv1, SDv3}, has come to the fore, demonstrating unparalleled capability in generating realistic visually novel content. Traditionally,
image generation with Stable Diffusion is guided by textual prompts \cite{TextInversion, dreambooth, FaceChain, SuDe, ImagineID}, which, while providing a high-level direction, lack the fine-grained instance information in crowd.

{
\textbf{Hence, refining the precision with which we can guide diffusion models has become a critical endeavor.} Existing methods such as DenseDiffusion~\cite{DenseDiffusion}, BoxDiff~\cite{BoxDiff}, and Free-Control~\cite{FreeControl} have made strides in directing the model's attention to specific regions of interest without requiring retraining, yet their ability to control dense instances remains inconsistent. In contrast, the proposed ControlNet~\cite{ControlNet} has shown promise, enabling the embedding of control signals directly into the Stable Diffusion framework via additional control maps. These maps, which can depict depth, edges, or density, allow users to precisely define the characteristics of the desired output image. However, \cite{WanJiaRej} has demonstrated that it remains challenging to use plain ControlNet for generating crowd images with instance-level annotations.
}

Considering these, our paper asserts there are two primary challenges hindering controlling images generation for crowd scenes: 
\textbf{1) efficient representation of dense objects in controlling map to instruct instance location; 2) injecting instance scale information to instruct instance size.}
Specifically, we demonstrate that ControlNet~\cite{ControlNet} is actually learning a projection from controlling space to the Stable Diffusion latent space. When we opt for an arbitrary representation of dense object, the disparate spaces incurs the hardness in controlling instance spatial location.  
Moreover, we empirically observe that the strong correlation between instance scale feature and spatial feature incurs the instance scales in generating images without providing additional scale guidance overfit to that in training set.

To navigate these challenges, we introduce DenseControl, whose framework is shown in Fig.~\ref{fig:framework}, that generates crowd images from user provided control over both the locations and scales of individual instances. DenseControl employs an Isolated Object Embedding (IOE) map as input to ControlNet~\cite{ControlNet}, representing each object with a distinct indicator positioned at the relevant instance location on the IOE map. Utilizing this IOE map as a control signal input simplifies the learning objective into a linear transformation, streamlining the projection into the Stable Diffusion latent space for enhanced spatial control. Additionally, we formulate an Implicit Scale Embedding (ISE) method to infuse scale information into the representation of instances, integrating the scale vector with the location indicators in the IOE map. Despite the complexity introduced by integrating scale embeddings, our novel Position Shortcut mechanism augments ControlNet’s original cross-attention functionality, significantly bolstering training efficiency.
Through comprehensive experimentation, we demonstrate that DenseControl achieves state-of-the-art in precisely controlling both the location and scale of instances in generated images. A comparative evaluation with ControlNet and various control map approaches underscores the superior efficacy of DenseControl. Moreover, we also show that based on precise control, DenseControl still shows strong textual instruction following capacity. Later, by leveraging the images synthesized by DenseControl, we significantly augment the performance of leading-edge crowd analysis models across scenarios of data scarcity, transfer learning, and weather adaptation.

In summary, our key contributions are:
\begin{itemize} 
    \item The introduction of DenseControl, a novel framework enabling precise instance-level manipulation in creating dense crowd images.
    \item The development of three innovative mechanisms: Isolated Object Embedding (IOE) for spatial control, Implicit Scale Embedding (ISE) for scale control without losing topological integrity, and a Position Shortcut for enhancing object projection. 
    \item Demonstrated superior performance in image synthesis quality and control precision, with practical applications that extend to data augmentation and transfer learning, showcasing DenseControl's broad utility. 
\end{itemize}

\section{Related Work}

\subsection{Crowd Analysis} 
\subsubsection{Crowd Counting} 
Crowd counting aims to estimate the number of people in images directly, primarily by generating density maps through a Gaussian-blurred annotation of heads, which neural networks then use for pixel-wise prediction. Key studies such as \cite{MCNN, CSRNet, BayesLoss, DENet, AdaCrowd, PFDNet, SASNet, wang2022counting} have successfully adopted this approach. The method effectively sums the density map values for a total count but falls short in distinguishing between individuals, as the density representations blend together, making discrete information extraction difficult.
\subsubsection{Crowd Localization} In contrast, crowd localization focuses on individually identifying and counting people, allowing for the extraction of spatial data on crowd constituents. Early attempts utilized object detection techniques \cite{DeTR}, but these faced challenges in dense crowds due to object overlap. Subsequent models like \cite{Tiny-Faces} refined detection by considering aspects such as resolution and context, whereas \cite{LSC-CNN, CrowdHat} enhanced performance through multi-scale analysis and unique feature exploitation.
Alternatively, methods like \cite{LiangFIDTM, Hantao} explore new frameworks bypassing traditional detection shortcomings, with strategies ranging from specialized density maps to novel distance mappings for improved localization. Despite advancements, pure density-based approaches exhibit limitations in precise localization.
Recent innovations like \cite{IIM, TopoCount} address these by delineating individuals as separate areas via segmentation, offering finer localization accuracy. Additionally, \cite{ScopedTeacher} proposes specific solutions for handling scale variations in instance-specific tasks. These developments illustrate the field's ongoing methodological evolution and diversification in crowd analysis.
\subsection{Controllable Image Diffusion Models}
The evolution of diffusion models~\cite{Diffusion} in image generation has been marked by pivotal advancements aimed at optimizing computational efficiency and enhancing image quality~\cite{VDM, CGDiff}. 
To enhance performance, latent diffusion models (LDM)~\cite{SDv1} propose transitioning the diffusion process into a latent space through Variational Autoencoders (VAE)~\cite{VAE}. 
Innovative solutions, such as Denoising Diffusion Implicit Models (DDIM)~\cite{DDIM}, aim to streamline this process by reducing the number of required steps to merely tens. Further developments, including \cite{eDiffI,ProgressiveDistillation,DistillationDiffusion}, have introduced methodologies to distill the lengthy step diffusion processes into significantly shorter ones, vastly improving efficiency.
\subsubsection{Text-Guided Diffusion Models}
The inception of text-guided diffusion models marked a significant milestone in generating high-quality, controllable images. Initial techniques, such as Classifier guidance~\cite{CGDiff}, leveraged classifiers to dictate the category of the generated image, with subsequent methods like \cite{CFG} introducing the addition of unconditional noise into the denoising process for enhanced control.

Efforts to attain finer control over generation led to the incorporation of CLIP-extracted~\cite{CLIP} text embeddings into the noise prediction network by Glide~\cite{Glide}, and Stable Diffusion~\cite{SDv1}, enabling arbitrary text-guided generation. Text Inversion~\cite{TextInversion} and Dreambooth~\cite{dreambooth} further personalized text-guided diffusion models. To improve adherence to textual instructions, researchers explored various strategies, including prompt adjustment, CLIP feature manipulation, and modification of cross-attention mechanisms~\cite{BlendedDiffusion, MakeAScene, Prompt2Prompt, DiffusionClip, Imagic, I2ITrans}. 
{Recently, some approaches have combined diffusion models with autoregressive prediction architectures to achieve high-quality text-to-content generation, including for images~\cite{li2024autoregressive}, videos~\cite{feng2025unified}, human motion~\cite{xiao2025motionstreamer}, and audio~\cite{yang2025generative}. Nevertheless, these approaches sometimes fall short in controlling the spatial distribution of generated content.}

\subsubsection{Spatial-Controllable Diffusion Models}
Building on the text-guided diffusion model framework, further research aimed at manipulating generated images' spatial attributes. Methods such as \cite{ConceptAblation, PaintExample} demonstrated how manipulating noise or enhancing classifier-free similarity could achieve concept manipulation or inpainting at a regional level. Systems like \cite{DAAM, BoxDiff, DenseDiffusion} have capitalised on augmenting or adjusting the UNet's~\cite{UNet} self-attention or cross-attention maps to influence spatial characteristics directly.

Innovations like FreeControl~\cite{FreeControl} introduced frameworks for decoupling spatial control into separate appearance and structure components or for maintaining consistency in spatial distribution according to a controlling map. MIGC~\cite{MIGC} addressed the challenge of multi-instance generation by breaking it down into subtasks for targeted enhancement. 
Based on these prior works, BoxDiff~\cite{BoxDiff} and DenseDiffusion~\cite{DenseDiffusion} propose to enhance the cross-attention maps between the region of interest with corresponding textual prompt.
ControlNet~\cite{ControlNet} proposes a unified framework, that can generate images with consistency spatial distribution with the any input controlling map.
However, despite these advancements, simultaneously controlling instance locations and scales in densely populated images remains a challenge yet to be fully addressed.

{
Recently, diffusion transformers~\cite{peebles2023scalable} or generative transformers~\cite{wang2025language} have emerged as powerful foundational base models. To endow them with spatial control capabilities, various methods have been proposed, including Generic Layout~\cite{wu2025hybrid}, RAGD~\cite{chen2025ragd}, ART~\cite{pu2025art}, and ControlAR~\cite{li2024controlar}. Despite achieving superior performance, transformer-based architectures impose a heavy computational burden, making them less efficient than the ControlNet series~\cite{ControlNet, qin2023unicontrol, li2024controlnetpp, wang2025synchronized} or agentic generation paradigms~\cite{Newton}.
}

\section{Methodology}
\begin{figure*}[t]
    \centering
    \includegraphics[width=1.0\textwidth]{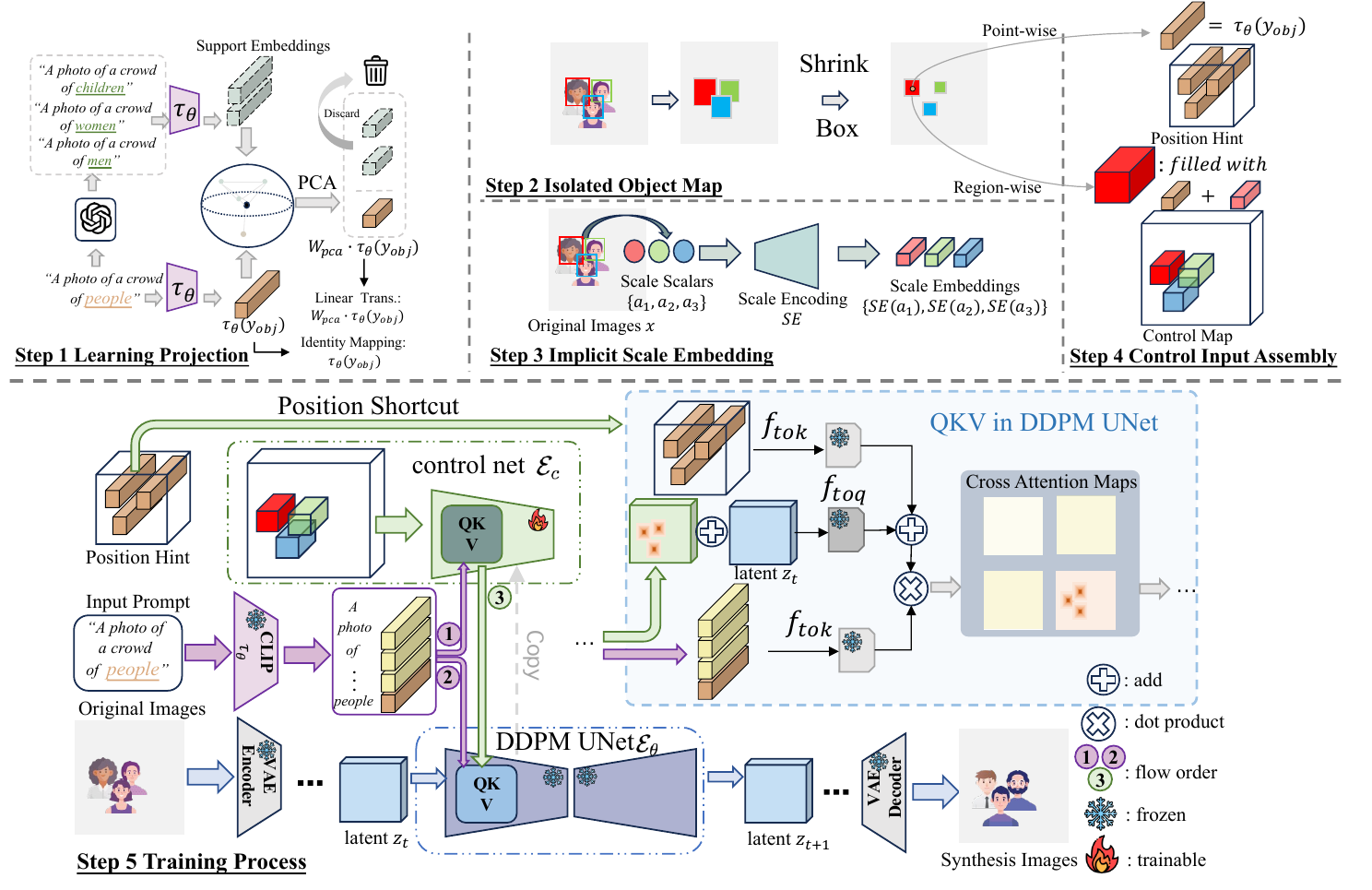}
    \caption{Pipeline of proposed DenseControl. The whole pipeline is divided into 5 steps, where the first 4 steps are finished before training. Step 1 and step 2 illustrate our proposed Isolated Object Embedding (IOE) to address Challenge 1 we raised in this paper, which is the difficulty of projection in controlling instance location. Step 3 and step 4 show the devised Implicit Scale Embedding (ISE), which is to cope with Challenge 2, injecting scale knowledge. Then, step 5 details our proposed Location Shortcut, which is to alleviate difficulty incurred by assembling IOE with ISE, and it also illustrates the detailed training flow.
    }
    \label{fig:pipeline}
\end{figure*}
In this section, commencing from preliminary on ControlNet~\cite{ControlNet}, we respectively identify and address two fundamental limitations of ControlNet in the context of generating crowd images, leading to our proposition of DenseControl.

\subsection{Preliminary on Stable Diffusion based ControlNet}
Stable diffusion~\cite{SDv1} processes involve generating images corresponding to user-provided text descriptions. This is achieved through a variational autoencoder (VAE)~\cite{VAE} that encodes an image \(\mathbf{x}\in\mathbb{R}^{3\times H\times W}\) into a latent representation \(\mathbf{z}\in\mathbb{R}^{4\times\frac{H}{8}\times\frac{W}{8}}\). Gaussian noise \(\epsilon\) is subsequently added in \(T\) steps until \(\mathbf{z}_T\) becomes purely noise. At each step \(t\), a denoising diffusion probabilistic model (DDPM) based on UNet~\cite{UNet}, \(\mathcal{E}_\theta\), predicts the added noise \(\epsilon_t\) from the noisy latent \(\mathbf{z}_{t-1}\), conditioned on time embedding and a textual prompt \(\mathbf{y}\), processed by a CLIP~\cite{CLIP} text encoder $\tau_{\theta}(\mathbf{y})\in\mathbb{R}^{d_c\times l_c}$, where $d_c$ is feature dimension and $l_c$ is the text length. Training of DDPM UNet follows
\begin{equation}\label{DDPMLoss}
    \mathcal{L}_{DDPM} = \mathbb{E}_{\mathbf{z},\mathbf{y},\epsilon \sim\mathcal{N}(0,1),t}\left \| \epsilon (t) - \epsilon _\theta[\mathbf{z}_t, t, \tau_{\theta}(\mathbf{y})] \right \|^2.
\end{equation}
To implement control, ControlNet, whose pipeline is illustrated by Fig.~\ref{fig:framework}, integrates an additional encoder, \(\mathcal{E}_c\), alongside an input control map \(\mathbf{h}\in\mathbb{R}^{C_d\times H_c\times W_c}\) to guide image generation, conditioned on textual embedding $\tau_\theta(\mathbf{y})$. 
During each forward pass, features from corresponding layers in the main and control UNets are summed before proceeding to the subsequent layer. 
This enables modified cross-attention mechanisms incorporating \(\mathbf{h}\), enhancing correspondence between generated image features and textual descriptions, as shown by Eq.~\ref{eq:ca}:
\begin{equation}\label{eq:ca}
    \mathbf{S}=\mbox{Softmax}(\frac{f_{toq}(\mathcal{E}_{\theta}(\mathbf{z}_t)+\mathcal{E}_c(\mathbf{h}))\cdot f_{tok}(\tau_{\theta}(\mathbf{y}))^\top }{\sqrt{d}}),
\end{equation}
where $f_{toq}\&f_{tok}$ denote the linear projections, $d$ is the hidden dimension.
This hybrid approach effectively combines stable diffusion with user-specified controls, anchored by the same loss function in Eq.~\ref{DDPMLoss}, to optimize only the control net \(\mathcal{E}_c\).

\subsection{Cumbersomeness of ControlNet}
The primary objective of the control net is to generate images in alignment with a control map $\mathbf{h}$, enhancing spatial consistency through targeted augmentation of the cross-attention mechanism with textual embeddings. This process results in a distribution of attention maps $\{\mathbf{s}_1, \ldots ,\mathbf{s}_{l_c}\}$, where each map coordinate $s^{xy}_{i}$ signifies the likelihood of representing a textual concept $\mathbf{y}_i$ at that location $(x, y)$. Consequently, the training objective of control network is designed to enhance attention in regions indicated by the control map while minimizing attention elsewhere, formalized as
\begin{align}\label{ControlNetObjective}
    \sum_{obj}\max f_{toq}\left( \epsilon ^{obj }_\theta(\mathbf{z}_t)+\mathcal{E}^{obj}_c(\mathbf{h})\right) \cdot f_{tok}\left( \tau_\theta(\mathbf{y})^{obj}\right) ^\top+\notag\\
    \min f_{toq}\left( \epsilon ^{obj }_\theta(\mathbf{z}_t)+\mathcal{E}^{\overline{obj}}_c(\mathbf{h})\right) \cdot f_{tok}\left( \tau_\theta(\mathbf{y})^{\overline{obj}}\right)^\top,
\end{align}
where $obj$ and $\overline{obj}$ elucidate guided by controlling map $\mathbf{h}$, the locations with or without the intended object, respectively. 
Hence, given the same control net $\mathcal{E}_c$, the final controlling performance greatly depends on the representation in controlling map $\mathbf{h}$.
To better control crowd image synthesis, we assert there are two main challenges.

\begin{figure}[t]
    \centering
    \includegraphics[width=0.45\textwidth]{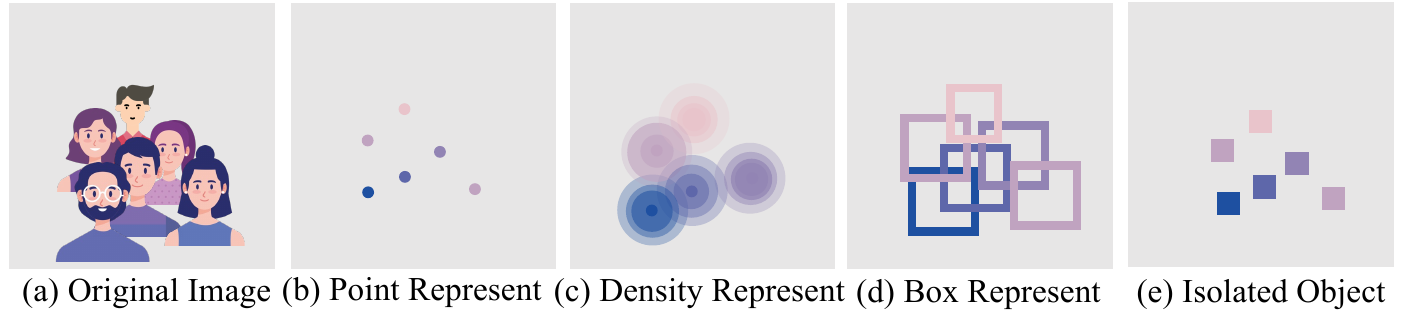}
    \caption{Three levels of object representation granularity in crowd images for the same image (a): point representation (b) is overly sparse, density representation (c) obscures instance boundaries, and bounding box representation (d) introduces topological errors due to overlap. Our proposed isolated box (e) strikes a balance among these granularities.}
    \label{fig:representFig}
\end{figure}

\textit{\textbf{Challenge 1\label{Challenge_1}: Efficient Representation to Instruct Instance Location}} \ 
To represent an instance, we can decouple into latent and spatial parts, which indicate the channel and width\&height in the controlling map respectively.
\underline{For the latent part}, existing $\mathbf{h}$ usually has a channel of 3 with arbitrary content, whose latent space is far away from stable diffusion. As a result, to optimize the objective of control net, namely Eq.~\ref{ControlNetObjective}, the $\mathcal{E}_c$ must learn to align the $\mathbf{h}$ space towards the stable diffusion space first, which requires an amount of training data and steps, and this is also supported by the empirical observation in ControlNet~\cite{ControlNet}.
\underline{For the spatial part}, as shown by Fig.~\ref{fig:representFig}, conventional representation formats (point, density, and box maps) cannot provide sufficient granularity. Concretely,
\begin{itemize}
    \item \emph{Point maps} are overly sparse, hindering effective control signal capture.
    \item \emph{Density maps} lack clear object boundaries, failing to convey precise instance-level detail.
    \item \emph{Box maps}, with overlapping bounding boxes, compromise the topological continuity of objects, obscuring individual instance guidance.
\end{itemize}
Moreover, while fine-grained maps (e.g., edge, depth, heat maps) offer detailed spatial cues, their granularity may excessively constrain the user ability to provide novel control signals, thereby limiting generalizability.
Thus, \emph{how to efficiently represent an instance in controlling map} hinders the precise instruction to location.

\textit{\textbf{Challenge 2: Injecting Instance Scale Information to Instruct Instance Size}} \ 
{
For crowd images, studies have demonstrated a strong correlation between instance scale, location, and background~\cite{ScopedTeacher}. Consequently, if explicit instance scale information is not provided during the training of ControlNet, this correlation may introduce bias, compromising the diversity of synthesized images.
However, the most common method for explicitly representing scale information is to use a bounding box. Yet as discussed earlier, this approach disrupts topological structures, rendering it a failure case.
Hence, \emph{how to inject instance scale knowledge without breaking original information} becomes the second challenge.}

\subsection{DenseControl}
We design DenseControl, illustrated as Fig.~\ref{fig:pipeline}. Specifically, we introduce Isolated Object Embedding (steps 1 and 2) to address challenge 1, Implicit Scale Embedding and Position Shortcuts (steps 3 and 4) for better incorporating scale information for challenge 2.

\subsubsection{Isolated Object Embedding (IOE)}
The IOE is composed of latent and spatial aspects.
\paragraph{Latent Projection Objective}
To tackle the projection difficulty incurred by distant latent spaces from controlling input of stable diffusion, we propose two variants of IOE to advance the projection of control net $\mathcal{E}_c(\mathbf{h})$ by learning an identity mapping and learning a linear transformation, respectively.

\textbf{Learning an Identity Mapping}
Considering the target space is the VAE latent space of stable diffusion, we can simplify the projection by directly using the corresponding latent representation for each instance.
However, the VAE latent representation is highly uncertain due to probabilistic sampling, making it unsuitable for our purposes here.
To cope with that, we can utilize the CLIP text embedding for the concept of \emph{people} to represent an instance. 
Formally, given $N$ instances represented by $\mathcal{I}=\{(x_i, y_i)\}_{i}^{N}$, where $(x_i, y_i)$ is the coordinate, the controlling map $\mathbf{h}\in\mathbb{R}^{d_c\times H\times W}$ can be generated by
\begin{equation}\label{IdentityMapping}
    \left\{
        \begin{array}{lcl}
        \mathbf{h}_{xy}=\tau_\theta(\mathbf{y}_{obj})&\mbox{, where }(x,y)\in \mathcal{I}, \\
        \mathbf{h}_{xy}=\mathbf{0}&\mbox{, where } (x,y)\notin  \mathcal{I}.
        \end{array}
    \right.
\end{equation}
The pre-trained stable diffusion effectively aligns the CLIP textual space with the VAE latent space, enhancing the similarity score in Eq.~\ref{ControlNetObjective}.
To this end, the projection objective for control net $\mathcal{E}_c$ can be an identity mapping.
   
\textbf{ Learning a Linear Transformation}
Despite the significant simplification provided by Eq.~\ref{IdentityMapping}, maintaining a control map with the same dimensionality as the CLIP text embedding is costly in terms of both hard disk and GPU storage.
Empirically, a single control map using double-precision floating-point numbers to guide the generation of a $512\times512$ image requires $1.5 Gb$ of space. Furthermore, as this control map serves as input for a trainable control network that necessitates gradient computation, even an $80 Gb$ A100 GPU cannot support training with a batch size of $1$.
As a result, we propose performing dimension reduction on the original CLIP text embedding. During this, it is crucial to guarantee that this dimension reduction should be some easy-to-learn transformations. 
Given this requirement, we employ principal component analysis (PCA)~\cite{PCA} to reduce the dimensionality of the embedding. However, in the task of crowd analysis, there is only a single object category, namely \emph{people}. A single category does not provide a sufficient subspace for PCA to operate effectively.
Therefore, we leverage a large language model (LLM) like GPT-4~\cite{GPT4} to generate text related to crowd scenes. These texts, combined with our category of interest, \emph{people}, are used to construct a PCA subspace and facilitate dimension reduction.
Formally, the controlling map $\mathbf{h}\in\mathbb{R}^{d_p\times H\times W}$ can be derived by
\begin{equation}\label{PCAMapping}
    \left\{
        \begin{array}{ll}
        \mathbf{h}_{xy}=\mathbf{W}_{pca}\cdot\tau_\theta(\mathbf{y}_{obj})&\mbox{, where }(x,y)\in \mathcal{I}, \\
        \mathbf{h}_{xy}=\mathbf{0}& \mbox{, where }(x,y)\notin  \mathcal{I},
        \end{array}
    \right.  
\end{equation}
$
\mbox{where }\mathbf{W}_{pca}=\mbox{PCA}[\tau_\theta(\mathbf{y}_{obj}),\tau_\theta(\mbox{LLM}(\mathbf{y}_{obj}))].
$

In this way, the final projection involves learning a linear transformation based on PCA, which presents a level of complexity that is still manageable for the control network to learn effectively.

\paragraph{Spatial Representation}
Due to the convolutional architecture of the control net, a controlling map with overly coarse granularity may result in excessively sparse gradients, complicating the optimization process. Furthermore, the granularity of the annotated bounding boxes is excessively coarse, as previously mentioned.
To attain an optimal granularity balance, we draw inspiration from IIM~\cite{IIM} and propose representing each object with a distinct bounding box, iteratively refined to eliminate overlaps by shrinking the boxes.
In this approach, we can generate a binary mask, denoted as $\mathbf{M}$, where each object is represented by a distinct foreground region derived from its correspondingly shrunk bounding box. Then, recalling the controlling map, for $i^{th}$ independent region, we can fill it up with the $\mathbf{h}_i=\mathbf{W}_{pca}\cdot\tau_\theta(\mathbf{y}_{obj})$.

\subsubsection{Implicit Scale Embedding}
To tackle Challenge 2,
which involves the incorporation of scale knowledge,
we embed the scalar representing the instance scale into an embedding vector and add it with the above PCA-reduced embedding.
To ensure that the embedding relationship across various scales is readily mappable and thus easily learnable by the control network, we employ cosine-sine embeddings to represent scale as a scalar input vector.
Specifically, complementing the instances set as $\mathcal{I}=\{(x_i, y_i), a_i\}_{i}^{N}$, where the $a_i$ denotes the instance scale area. The controlling map $\mathbf{h}\in\mathbb{R}^{d_p\times H\times W}$ is generated by
\begin{equation}\label{withScaleEmbedding}
    \left\{
        \begin{array}{ll}
        \mathbf{h}_{xy}=\mathbf{W}_{pca}\cdot\tau_\theta(\mathbf{y}_{obj})+\mbox{SE}(a)&\mbox{, where }(x,y)\in \mathcal{I}, \\
        \mathbf{h}_{xy}=\mathbf{0}& \mbox{, where }(x,y)\notin  \mathcal{I},
        \end{array}
    \right.  
\end{equation}
\begin{equation}\label{ScaleEmbedding}
    \left\{
        \begin{array}{ll}
        \mbox{SE}(a)_{2i}=\sin(\frac{a}{10000^{2i/d_p}}), \\
        \mbox{SE}(a)_{2i+1}=\cos(\frac{a}{10000^{2i/d_p}}).
        \end{array}
    \right.  
\end{equation}

\textbf{Position Shortcut}
Through Eq.~\ref{withScaleEmbedding}, we incorporate instance-wise scale information into the controlling map. However, it increases the complexity of projecting the embedding into our target space. To mitigate this challenge, we propose enhancing the cross-attention mechanism for the object of interest by introducing explicit location information through a Position Shortcut.
Specifically, referring to the control map in Eq.~\ref{IdentityMapping}, which is composed of the original CLIP text embedding and contains explicit location information, we can optimize the use of this location information by reducing its resolution to match that of the VAE latent space. To achieve this, we define the location hint as
 $\widehat{\mathbf{h}}$, sharing the same spatial dimension with the VAE latent that is $\frac{H}{8}\times\frac{W}{8}$. It is composed of
\begin{equation}\label{ResidualConnection}
    \left\{
        \begin{array}{lcl}
        \widehat{\mathbf{h}}_{{x}' {y}' }=\tau_\theta(\mathbf{y}_{obj})&\mbox{, where }({x}' \times 8,{y}'\times8 )\in \mathcal{I}, \\
        \widehat{\mathbf{h}}_{{x}' {y}' }=\mathbf{0}&\mbox{, where }({x}' \times 8,{y}'\times8 )\notin \mathcal{I}.
        \end{array}
    \right.
\end{equation}
Then, the modified cross-attention scores can be computed by
\begin{equation}\label{ModifiedCA}
    \begin{aligned}
    &\mathbf{S} =  \text{Softmax}\left(\frac{(\mathbf{Q}+\mathbf{R})\cdot \mathbf{K}^\top }{\sqrt{d}}\right), \text{ where } \mathbf{K} = f_{tok}(\tau_{\theta}(\mathbf{y})), \\
    & \mathbf{R} = f_{tok}(\widehat{\mathbf{h}} ), \mathbf{Q} = f_{toq}(\mathbf{z}_t+\mathcal{E}_c(\mathbf{W}_{pca}\cdot\tau_\theta(\mathbf{y}_{obj})+\text{SE}(a))).
    \end{aligned}
\end{equation}
In Eq.~\ref{ModifiedCA}, the final cross-attention score to the $\mathbf{y}_{obj}$ is enhanced by the Position Shortcut of $\mathbf{R}$. 
Hence, incorporating it with the controlling map could achieve easier projection with introducing additional scale information.

{
\subsection{DenseControl User Guidance}
To facilitate practical adoption of DenseControl for instance-level dense crowd image synthesis, we provide clear, actionable guidelines tailored to user-specific control signal design and scenario requirements. These guidelines ensure reproducibility and maximize the framework’s utility across real-world applications.

\paragraph{Instance Scale Customization}
DenseControl supports full flexibility in per-instance scale definition—no uniformity constraints are imposed. 
Stability for extreme scale combinations is guaranteed by training on datasets with naturally diverse scale distributions. 
No additional hyperparameter tuning is required for scale customization—each instance’s scale is independently encoded via the Implicit Scale Embedding (ISE) module.

\paragraph{Density Configuration Recommendations}
DenseControl imposes no fixed density requirements, enabling free design of crowd layouts. 
Rare edge cases (e.g., ultra-dense layouts with ultra-large instances, or ultra-sparse layouts with ultra-tiny instances) may cause minor performance degradation due to limited real-world data coverage. However, core controllability (instance location/scale accuracy) remains uncompromised.

\paragraph{Guaranteed Background in Empty Regions}
Regions left empty in the control map \textit{exclusively generate background} (no unintended crowd instances). This behavior is enforced by the IOE module’s design:
\begin{itemize}
    \item The IOE map establishes a one-to-one correspondence between control signals (instance locations/scales) and generated objects.
    \item Empty regions in the IOE map provide no activation to the diffusion model, which defaults to synthesizing background content (e.g., ``empty sky'' for outdoor scenes, ``empty floor'' for indoor scenes).
\end{itemize}
This guarantee is critical for scenarios requiring precise crowd exclusion (e.g., ``crowd only in the stadium stands, not on the field'').
}

\section{Experiment}
\subsection{Experimental Setting}
\subsubsection{Datasets}

{

\paragraph{Test-Set Split Strategy}
To comprehensively validate DenseControl’s performance across diverse crowd scenarios and ensure experimental reproducibility, we design a multi-dimensional test-set splitting strategy based on two core attributes of crowd scenes: \textit{instance scale} and \textit{crowd density}. This strategy aligns with the statistical distribution of real-world crowd datasets and covers both common and edge-case scenarios.

There are six mainstream datasets for crowd analysis used in this paper, including SHHA~\cite{MCNN}, SHHB~\cite{MCNN}, SHRGBD~\cite{SHRGBD}, NWPU~\cite{NWPU}, JHU~\cite{JHU}, and QNRF~\cite{QNRF}. 
All datasets provide instance-level head bounding box annotations (officially~\cite{SHRGBD,NWPU,JHU} or from Wang, et al~\cite{ScaleBench}), which we use to quantify scale and density metrics.

With above dataset, we define two objective, annotation-based metrics to avoid subjective bias:
\begin{enumerate}
    \item \textit{Instance Scale}: The \textit{average area of head bounding boxes} per image (in pixels$^2$), denoted $S_{\text{avg}}$. This metric directly reflects the physical size of individuals (e.g., foreground heads have larger $S_{\text{avg}}$ than background heads).
    \item \textit{Crowd Density}: The \textit{total number of instances} per image, denoted $C_{\text{total}}$. This metric captures crowd sparsity/density (e.g., $C_{\text{total}} < 20$ for small groups, $C_{\text{total}} > 100$ for packed crowds).
\end{enumerate}

The test set is partitioned into four mutually exclusive categories (BigFew, SmallFew, BigMany, SmallMany) via a \textit{data-driven median split}.
}

\begin{table*}[]
\centering
\caption{Comparison with other controllable image generators. The smaller the values, the better the performance. $^*$ means an $80Gb$ GPU cannot support the memory allocation.}

\label{tab:CompOther}
\renewcommand{\arraystretch}{1.35}{
\resizebox{.95\textwidth}{!}{
\begin{tabular}{c|cccc|cccc|cccc|cccc}
\whline
Testing Set                  & \multicolumn{4}{c|}{BigFew}                                                                                                     & \multicolumn{4}{c|}{SmallFew}                                                                                                      & \multicolumn{4}{c|}{BigMany}                                                                                                   & \multicolumn{4}{c}{SmallMany}                                                                                                      \\ \hline
Method              & \multicolumn{1}{c|}{$FID$$^\downarrow $}             & \multicolumn{1}{c|}{$MAE$$^\downarrow $}           & \multicolumn{1}{c|}{$rMSE$$^\downarrow $}           & $NAE$$^\downarrow $           & \multicolumn{1}{c|}{$FID$}             & \multicolumn{1}{c|}{$MAE$}            & \multicolumn{1}{c|}{$rMSE$}            & $NAE$            & \multicolumn{1}{c|}{$FID$}             & \multicolumn{1}{c|}{$MAE$}           & \multicolumn{1}{c|}{$rMSE$}          & $NAE$           & \multicolumn{1}{c|}{$FID$}             & \multicolumn{1}{c|}{$MAE$}             & \multicolumn{1}{c|}{$rMSE$}            & $NAE$           \\ \whline
BoxDiff~\cite{BoxDiff}             & \multicolumn{1}{c|}{45.71}          & \multicolumn{1}{c|}{300.23}        & \multicolumn{1}{c|}{371.43}         & 33.97         & \multicolumn{1}{c|}{59.35}          & \multicolumn{1}{c|}{254.71}         & \multicolumn{1}{c|}{312.7}           & 20.35          & \multicolumn{4}{c|}{\textit{out of memory}$^*$}                                                                                             & \multicolumn{4}{c}{\textit{out of memory}}                                                                                                  \\ \hline
DenseDiffusion~\cite{DenseDiffusion}        & \multicolumn{1}{c|}{\underline{38.53}}          & \multicolumn{1}{c|}{318.49}        & \multicolumn{1}{c|}{540.48}         & 38.09         & \multicolumn{1}{c|}{45.12}          & \multicolumn{1}{c|}{142.55}         & \multicolumn{1}{c|}{\textbf{294.66}}          & 11.42          & \multicolumn{4}{c|}{\textit{out of memory}}                                                                                             & \multicolumn{4}{c}{\textit{out of memory}}                                                                                                  \\ \hline
MIGC~\cite{MIGC}                  & \multicolumn{1}{c|}{43.10}          & \multicolumn{1}{c|}{\underline{77.20}}          & \multicolumn{1}{c|}{\underline{197.54}}         & \underline{9.67}          & \multicolumn{1}{c|}{\underline{41.74}}          & \multicolumn{1}{c|}{\underline{138.67}}         & \multicolumn{1}{c|}{324.69}          & 10.93          & \multicolumn{4}{c|}{\textit{out of memory}}                                                                                             & \multicolumn{4}{c}{\textit{out of memory}}                                                                                                  \\ \hline
FreeControl~\cite{FreeControl}           & \multicolumn{1}{c|}{54.39}          & \multicolumn{1}{c|}{607.04}        & \multicolumn{1}{c|}{668.96}         & 71.83         & \multicolumn{1}{c|}{60.52}          & \multicolumn{1}{c|}{665.57}         & \multicolumn{1}{c|}{727.78}          & 59.42          & \multicolumn{1}{c|}{46.72}          & \multicolumn{1}{c|}{657.18}        & \multicolumn{1}{c|}{727.79}        & 12.65         & \multicolumn{1}{c|}{55.12}          & \multicolumn{1}{c|}{674.01}          & \multicolumn{1}{c|}{765.68}          & 14.53         \\ \hline
\textbf{DenseControl (ours)} & \multicolumn{1}{c|}{\textbf{33.17}} & \multicolumn{1}{c|}{\textbf{5.34}} & \multicolumn{1}{c|}{\textbf{25.64}} & \textbf{0.72} & \multicolumn{1}{c|}{\textbf{39.58}} & \multicolumn{1}{c|}{\textbf{82.36}} & \multicolumn{1}{c|}{\underline{300.23}} & \textbf{10.38} & \multicolumn{1}{c|}{\textbf{27.86}} & \multicolumn{1}{c|}{\textbf{2.93}} & \multicolumn{1}{c|}{\textbf{5.37}} & \textbf{0.04} & \multicolumn{1}{c|}{\textbf{35.47}} & \multicolumn{1}{c|}{\textbf{194.97}} & \multicolumn{1}{c|}{\textbf{345.53}} & \textbf{4.46} \\ \whline
\end{tabular}
}}
\end{table*}

\subsubsection{Evaluation Metrics}
We assess the performance of the proposed DenseControl method on two primary criteria: 1) the quality of the data it generates and  2) its efficacy in real-world crowd localization applications.
For criteria 1), we utilize four metrics following other generation works~\cite{SDv1, SDv3, ControlNet}. \textbf{Fréchet Inception Distance} ($FID$)~\cite{FID} is to compare the similarity between the generated image and ground-truth image. \textbf{Human Preference} ($HP$)~\cite{ControlNet} compares images generated by different ControlNet representations but conditioned by the same controlling signal. \textbf{Average Precision 50} ($AP@50$) measures the position and scale controlling precision by comparing boxes in the controlling signal with that by human-annotated. \textbf{Attribute Retrieval} ($AR$) judges whether the attributes included in the instruction are generated.
For criteria 2), we utilize \textbf{F1-measure} ($F_1$), \textbf{Precision} ($Pre.$), and \textbf{Recall} ($Rec.$) to measure the localization performance, while \textbf{Mean Absolute Error} ($MAE$), \textbf{rooted Mean Square Error} ($rMSE$), and \textbf{Normed Absolute Error} ($NAE$) are to measure counting performance, following previous crowd analysis works~\cite{TopoCount,IIM}.

\begin{figure}
    \centering
    \includegraphics[width=.48\textwidth]{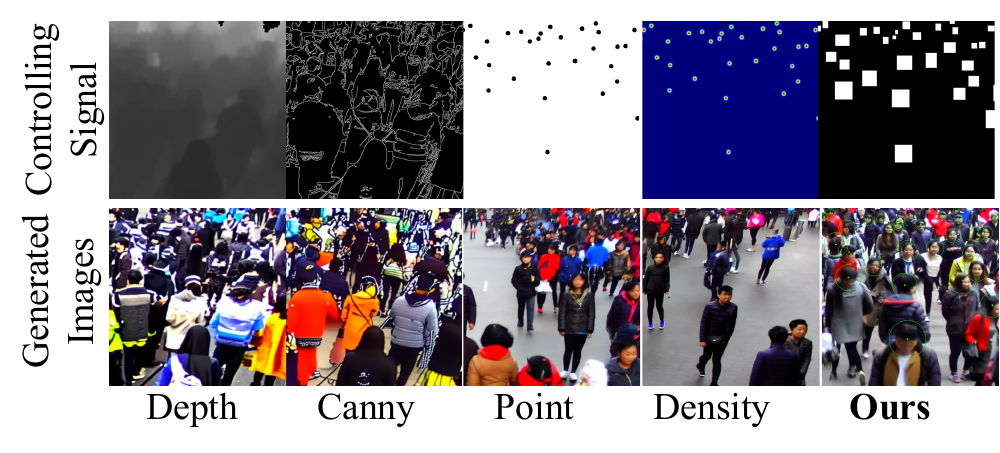}
    \caption{Visulization results of comparison of different controlling signals, including depth, canny, point, and density. 
    }
    \label{fig:Modality}
\end{figure}

\subsection{Comparison with Other Controllable Image Generators}\label{sec:compare_sota}

\subsubsection{Setting}  
We measure the quality of generated images over both different instance scales (\emph{Big/Small}) and counts (\emph{Many/Few}).
Specifically, we split the test sets of aforementioned datasets into the following four categories: BigFew, SmallFew, BigMany, and SmallMany.
Among them, BigMany indicates instances on a large scale and a high quantity.

\subsubsection{Result Analysis}
\tableautorefname~\ref{tab:CompOther} shows our method outperforming state-of-the-art generative models, including BoxDiff~\cite{BoxDiff}, DenseDiffusion~\cite{DenseDiffusion}, MIGC~\cite{MIGC}, and FreeControl~\cite{FreeControl}, over different scales and counts under previous defined four categories.
We observe two primary insights:
\begin{itemize}
\item 
\textit{Superior generative performance in FID:} Across all scenarios, our approach yields the lowest $FID$ scores, improving over the runner-up by non-trivial improvement, indicating unmatched result quality.
\item
\textit{Adaptability and efficiency under constraints:} While each method has its preferred application conditions, ours proves especially effective in large-scale, high-density settings, overcoming memory limitations that hinder BoxDiff, DenseDiffusion, and MIGC. It also consistently delivers the most precise head counts, showing its potential for crowded scene synthesis despite slightly underperforming in the SmallFew rMSE metric.
\end{itemize}
\begin{figure*}[t]
    \centering
    \includegraphics[width=1.\textwidth]{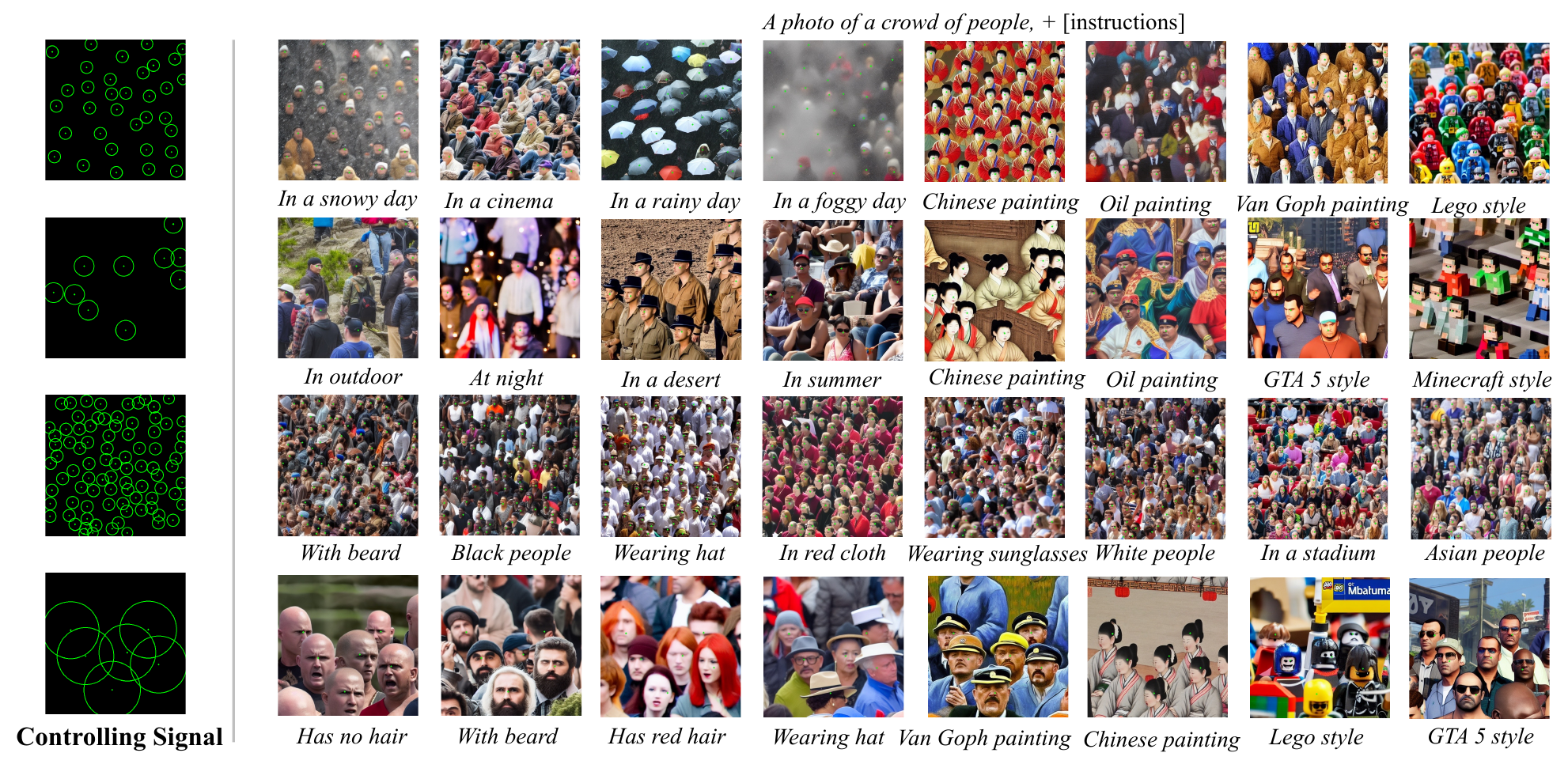}
    \caption{Controlling with any textual instructions in generating crowd images. Given a kind of controlling signal, including instances locations and scales, we can generate corresponding crowd images following the additionally provided textual instructions. Best view in color and zoom.}
    \label{fig:enter-label}
\end{figure*}

\begin{table}[t]
\centering
\caption{Quantitative comparison with other kinds of representation formats as the control signal to generate crowd images.}

\label{tab:CompRepresent}
\renewcommand{\arraystretch}{1.35}{
\resizebox{.50\textwidth}{!}{
\begin{tabular}{c|cc|ccc|c}
\whline
                            & \multicolumn{2}{c|}{Overall Quality}                                                         & \multicolumn{3}{c|}{Global Control}                                                                                                          & Instance Control   \\ \cline{2-7} 
\multirow{-2}{*}{Represent} & \multicolumn{1}{c|}{$FID^\downarrow $}                      & $HP^\downarrow $ & \multicolumn{1}{c|}{$MAE^\downarrow $}                     & \multicolumn{1}{c|}{$rMSE^\downarrow $}                     & $NAE^\downarrow $ & $AP@50^\uparrow $ \\ \whline
Canny                       & \multicolumn{1}{c|}{125.86}                                 & 4.20                           & \multicolumn{1}{c|}{10.54}                                 & \multicolumn{1}{c|}{15.41}                                  & 0.49              & 14.96              \\ \hline
Depth                       & \multicolumn{1}{c|}{71.68}                                  & 4.81                           & \multicolumn{1}{c|}{28.53}                                 & \multicolumn{1}{c|}{53.56}                                  & 1.81              & 20.04              \\ \hline
Point                       & \multicolumn{1}{c|}{23.05}                                 & 3.19                           & \multicolumn{1}{c|}{37.93}                                 & \multicolumn{1}{c|}{46.89}                                  & 3.4               & 2.69               \\ \hline
Density                     & \multicolumn{1}{c|}{21.81}                                 & 2.82                           & \multicolumn{1}{c|}{26.38}                                 & \multicolumn{1}{c|}{31.78}                                  & 2.36              & 3.22               \\ \hline
\rowcolor[HTML]{EFEFEF} 
DenseControl (ours)         & \multicolumn{1}{c|}{\cellcolor[HTML]{EFEFEF}\textbf{4.98}} & \textbf{1.36}                  & \multicolumn{1}{c|}{\cellcolor[HTML]{EFEFEF}\textbf{4.42}} & \multicolumn{1}{c|}{\cellcolor[HTML]{EFEFEF}\textbf{11.72}} & \textbf{0.28}     & \textbf{32.94}  \\
\whline
\end{tabular}
}}
\end{table}

\subsection{Comparison among Different Representations}\label{sec:represent}

\subsubsection{Setting} 
Considering our baseline method ControlNet being adaptable with variant controlling representation formats, in this section, we train ControlNet
on SHRGBD~\cite{SHRGBD} with Depth, Canny, Point, and Density map~\cite{WanJiaRej} as representation formats to control image generation process.
\subsubsection{Result Analysis}
In~\figurename~\ref{fig:Modality}, it is obvious that our proposed representation as the control signal could generate clearer and more consistent images compared with other formats (Depth, Canny, Point, and Density).
Our quantitative assessment in \tablename~\ref{tab:CompRepresent}, solidifies this finding across three metrics:
\begin{itemize}
\item 
\textit{Overall Quality:} Our representation format achieves the best $FID$ score, an improvement of approximately 16 points over the nearest competitor, correlating with higher Human Preference ratings.
\item \textit{Global Control:} Our method, emphasizing accurate instance counts, showcases significant progress by exceeding the second-best performance by 6.12 points on $MAE$.
\item  \textit{Instance Control:} When comparing controlled boxes against human annotations, DenseControl achieves notable improvements over other methods, with a 12.9\% increase in $AP@50$.
\end{itemize}

{
\subsubsection{Why not shrinking density or scaling point?}
We chose to shrink the bounding box rather than using points or density for the following reasons:
\begin{enumerate}
    \item \textbf{Bounding box} (our solution): The bounding box is an effective representation for indicating the precise location and spatial extent of an object. By shrinking the bounding box, we maintain a clear spatial representation of the object while avoiding overlap with adjacent instances.
    \item \textbf{Points}: Using points alone is not ideal for dense object distributions, as they tend to be too sparse and fail to provide sufficient spatial cues. Points do not explicitly represent the extent or boundaries of an object, which would complicate spatial control.
    \item \textbf{Density}: Using density for shrinking is problematic because pixel values vary within the same object region. This inconsistency could lead to blurred boundaries and an imprecise representation of object locations. Shrinking with density would thus blur the distinctions between objects, violating the requirement for clear object boundaries.
\end{enumerate}
Thus, shrinking the bounding box provides a reliable and straightforward method to preserve object independence while maintaining clear boundaries, which is crucial for subsequent operations in the model.
}

\subsection{Performance Evaluation on the Instruction Following} \label{sec:instruction}

\subsubsection{Setting} 

We use different textual instructions for crowd image generation to observe whether the method could understand and execute the instructions, based on our proposed spatial and scale control.
In this paper, we utilize three different types of instructions, indicating background, instance, and scale. For each type instructions, we utilize GPT-4 to generate 20 different instructions. 
During inference, we add these instructions with original prompt to generate images.

\begin{table}[t]
\centering
\caption{The performance on instruction following in generating crowd images.}

\label{tab:Instruction}
\renewcommand{\arraystretch}{1.35}{
\resizebox{.45\textwidth}{!}{
\begin{tabular}{c|c|c|cc}
\whline
\multirow{2}{*}{Instruction Type} & Generation           & Detection               & \multicolumn{2}{c}{Counting}                              \\ \cline{2-5} 
                                  & $AR$ (\%)$^\uparrow$ & $AP@50$ (\%)$^\uparrow$ & \multicolumn{1}{c|}{$MAE^\downarrow$} & $rMSE^\downarrow$ \\ \whline
Background                        & 93.21                & 9.91                    & \multicolumn{1}{c|}{20.48}            & 36.39             \\ \hline
Style                             & 100.00               & 40.56                   & \multicolumn{1}{c|}{4.05}             & 7.11              \\ \hline
Instance                          & 66.83                & 27.35                   & \multicolumn{1}{c|}{5.89}             & 12.12             \\ \whline
\end{tabular}
}}
\end{table}
\subsubsection{Result Analysis}
Fig.~\ref{fig:enter-label} showcases our method's adherence to provided instructions with high consistency. Our quantitative evaluation (\tableautorefname~\ref{tab:Instruction}) reveals:
\begin{itemize}
    \item 
\textit{Background and Style Attributes:} We achieve remarkable accuracy, reaching $93.21\%$ for background attributes and a perfect score of $100\%$ for style, underscoring robust control in image generation.
\item \textit{Instance Generation:} Despite challenges in instance diversity, our accuracy stands at $66\%$. Lower performance is achieved with background type (e.g., \emph{in a rainy day}). According to feedback from human annotators, this should be blamed on umbrella occlusions.
\end{itemize}

\subsection{Data Augmentation under Data Scarcity Scenario}\label{sec:DataScarcity}
\begin{table*}[]
\centering
\caption{Crowd localization and counting results when augmenting diffusion models generated images under data scarcity scenario.}

\label{tab:FewShot}
\renewcommand{\arraystretch}{1.35}{
\resizebox{.85\textwidth}{!}{
\begin{tabular}{c|cc|cc|cc}
\whline
\multirow{2}{*}{Method}                              & \multicolumn{2}{c|}{SHHA}                                                           & \multicolumn{2}{c|}{NWPU}                                                           & \multicolumn{2}{c}{QNRF}                                                            \\ \cline{2-7} 
                       & \multicolumn{1}{c|}{$\mathbf{F_1}$ / $Pre.$ / $Rec.$ (\%)$^\uparrow$}                        & $\mathbf{MAE}$ / $rMSE$$^\downarrow$ & \multicolumn{1}{c|}{$\mathbf{F_1}$ / $Pre.$ / $Rec.$ (\%)$^\uparrow$}                        & $\mathbf{MAE}$ / $rMSE$$^\downarrow$ & \multicolumn{1}{c|}{$\mathbf{F_1}$ / $Pre.$ / $Rec.$ (\%)$^\uparrow$}                        & $\mathbf{MAE}$ / $rMSE$$^\downarrow$ \\ \whline
IIM                           & \multicolumn{1}{c|}{63.0 / 63.3 / 62.7}                         & 109.2 / 185.9     & \multicolumn{1}{c|}{37.2 / 41.7 / 33.5}                         & 353.9 / 900.3     & \multicolumn{1}{c|}{44.7 / 57.8 / 36.4}                         & 461.4 / 1125.7    \\  \hline
\rowcolor[HTML]{EFEFEF} 
\textbf{IIM+DenseControl}     & \multicolumn{1}{c|}{\cellcolor[HTML]{EFEFEF}65.1 / 67.1 / 63.1} & 111.3 / 189.1     & \multicolumn{1}{c|}{\cellcolor[HTML]{EFEFEF}42.4 / 46.7 / 38.8} & 303.4 / 785.7     & \multicolumn{1}{c|}{\cellcolor[HTML]{EFEFEF}47.1 / 59.0 / 39.1} & 422.2 / 1032.7    \\ \whline
STEERER                       & \multicolumn{1}{c|}{61.3 / 50.2 / 78.7}                         & 91.8 / 131.3      & \multicolumn{1}{c|}{61.5 / 66.4 / 57.3}                         & 223.5 / 790.9     & \multicolumn{1}{c|}{61.6 / 58.3 / 65.3}                         & 168.2 / 271.0     \\ \hline
\rowcolor[HTML]{EFEFEF} 
\textbf{STEERER+DenseControl} & \multicolumn{1}{c|}{\cellcolor[HTML]{EFEFEF}63.7 / 55.1 / 75.4} & 99.3 / 145.8      & \multicolumn{1}{c|}{\cellcolor[HTML]{EFEFEF}60.5 / 70.5 / 53.0} & 180.4 / 508.8     & \multicolumn{1}{c|}{\cellcolor[HTML]{EFEFEF}62.1 / 59.7 / 64.7} & 195.0 / 328.9     \\ \whline
\end{tabular}
}}
\end{table*}
\subsubsection{Setting}
We randomly sample about $10\%$, $5\%$, and $1\%$ training samples from the training set of SHHA, QNRF, and NWPU.
We conduct experiments on crowd localization models: IIM and STEERER, with or without training on the synthesized images from DenseControl as data augmentation.

\subsubsection{Result Analysis} 
As evidenced by \tableautorefname~\ref{tab:FewShot}, our method consistently outperforms the IIM algorithm and the Density augmentation method across all datasets in terms of the F1Score, with enhancements of $2\%$, $5\%$, and $3\%$ and marginal gains of $0.6\%$, $0.3\%$, and $0.5\%$, respectively. While it shows variegated results against the STEERRER method—progressing on the SNHA and QNRF yet receding on the NWPU, likely due to noise in the images—our approach emerges as a promising solution for few-shot learning challenges.
However, similar to \tableautorefname~\ref{tab:CompOther}, we observe oscillations in MAE and rMSE metrics, indicating count inaccuracies that could affect model training. Future explorations will delve into fusing noise-resilient algorithms to mitigate this, highlighting an exciting direction for our subsequent work.

\subsection{Ablation Studies}\label{sec:ablation}
\begin{table}[]
\centering
\caption{Ablation studies of DenseControl, which are conducted on augmenting SHHA crowd localization environment.}

\label{tab:Ablation}
\renewcommand{\arraystretch}{1.35}{
\resizebox{.5\textwidth}{!}{
\begin{tabular}{c|c|c|c|c|c|c}
\whline
Module & Isolated Box & IOE & ISE & PS & $\mathbf{F_1}$ / $Pre.$ / $Rec.$ (\%)$^\uparrow$    & $\mathbf{MAE}$ / rMSE$^\downarrow$                        \\ \whline
Point RGB   &     &     &       &         & 60.4 / 55.4 / 66.5 & 152.2 / 259.6                      \\ \hline
IB+RGB    &  \ding{51}    &     &       &         & 63.9 / 65.0 / 62.7 & 112.5 / 194.1                      \\ \hline
Point+IOE   &     & \ding{51}    &       &         & 60.4 / 56.0 / 65.5 & 152.2 / 238.2                      \\ \hline
IB+IOE  &\ding{51}     &\ding{51}     &       &         &64.4 / 65.5 / 63.4  & 112.7 / 188.0                      \\ \hline
IB+ISE   & \ding{51}    &     & \ding{51}      &         & 64.2 / 70.2 / 59.2 & 116.8 / 218.2                      \\ \hline
Point+IOE+ISE  &     & \ding{51}    & \ding{51}      &         & 63.1 / 61.1 / 65.3 & 125.5 / 218.1                      \\ \hline
IB+IOE+ISE   & \ding{51}    & \ding{51}    & \ding{51}      &         &64.7 / 64.7 / 64.8  & 112.1 / 193.8                         \\ \hline
\cellcolor[HTML]{EFEFEF}DenseControl   &\cellcolor[HTML]{EFEFEF}\ding{51}     & \cellcolor[HTML]{EFEFEF}\ding{51}    & \cellcolor[HTML]{EFEFEF}\ding{51}      &\cellcolor[HTML]{EFEFEF}\ding{51}         & \cellcolor[HTML]{EFEFEF}65.1 / 67.1 / 63.1 &\cellcolor[HTML]{EFEFEF} 111.3 / 189.1                     \\ \whline
\end{tabular}
}}
\end{table}
\begin{table*}[t]
\centering
\caption{Crowd localization and counting results when augmenting diffusion models generated images under transfer-dataset scenario.}

\label{tab:Transfer}
\renewcommand{\arraystretch}{1.35}{
\resizebox{.85\textwidth}{!}{
\begin{tabular}{c|cc|cc|cc}
\whline
\multirow{2}{*}{Method} & \multicolumn{2}{c|}{SHHA$\to $SHHB}                  & \multicolumn{2}{c|}{SHHA$\to $QNRF}                    & \multicolumn{2}{c}{SHHA$\to $NWPU}                          \\ \cline{2-7} 
                        & \multicolumn{1}{c|}{$\mathbf{F_1}$ / $Pre.$ / $Rec.$ (\%)$^\uparrow$}     & $\mathbf{MAE}$ / $rMSE$ $^\downarrow$  & \multicolumn{1}{c|}{$\mathbf{F_1}$ / $Pre.$ / $Rec.$ (\%)$^\uparrow$}     & $\mathbf{MAE}$ / $rMSE$ $^\downarrow$    & \multicolumn{1}{c|}{$\mathbf{F_1}$ / $Pre.$ / $Rec.$ (\%)$^\uparrow$}         & $\mathbf{MAE}$ / $rMSE$ $^\downarrow$     \\ \whline
IIM                     & \multicolumn{1}{c|}{61.6 / 57.1 / 67.0} & 39.4 / 50.8 & \multicolumn{1}{c|}{50.8 / 53.9 / 48.1} & 356.9 / 627.0 & \multicolumn{1}{c|}{47.4 / 50.2 / 44.8}     & 221.7 / 734.0  \\
IIM+DenseControl        & \multicolumn{1}{c|}{67.0 / 64.7 / 69.5} & 29.3 / 41.1 & \multicolumn{1}{c|}{54.7 / 63.3 / 48.2} & 298.8 / 569.6 & \multicolumn{1}{c|}{51.3 / 54.6 / 48.4}     & 195.5 / 637.3  \\ \hline
STEERER                 & \multicolumn{1}{c|}{58.1 / 49.6 / 69.9} & 38.6 / 54.5 & \multicolumn{1}{c|}{40.0 / 27.2 / 75.9} & 244.8 / 663.8 & \multicolumn{1}{c|}{33.6  /  22.0  /  71.2} & 466.8 / 1087.8 \\
STEERER+DenseControl    & \multicolumn{1}{c|}{64.9 / 61.3 / 64.2} & 33.4 / 50.8 & \multicolumn{1}{c|}{45.5 / 33.4 / 71.4} & 288.6 / 537.9 & \multicolumn{1}{c|}{39.2  /  28.2 / 64.4}  & 506.8 / 1046.1 \\ \whline
\end{tabular}
}}
\end{table*}
\begin{table*}[t]
\centering
\caption{Crowd localization and counting results when transferring sunny scenes trained model to bad weather scenes.}

\label{tab:Weather}
\renewcommand{\arraystretch}{1.65}{
\resizebox{.9\textwidth}{!}{
\begin{tabular}{c|cc|cc|cc|cc}
\whline
\multirow{2}{*}{Setting} & \multicolumn{2}{c|}{Sunny$\to $Sunny (InD)}                    & \multicolumn{2}{c|}{Sunny$\to $Rainy}                         & \multicolumn{2}{c|}{Sunny$\to $Snowy}                         & \multicolumn{2}{c}{Sunny$\to $Foggy}                          \\ \cline{2-9} 
                         & \multicolumn{1}{c|}{$\mathbf{F_1}$ / $Pre.$ / $Rec.$ (\%)$^\uparrow$} & $\mathbf{MAE}$ / $rMSE$ $^\downarrow$    & \multicolumn{1}{c|}{$\mathbf{F_1}$ / $Pre.$ / $Rec.$ (\%)$^\uparrow$} & $\mathbf{MAE}$ / $rMSE$ $^\downarrow$   & \multicolumn{1}{c|}{$\mathbf{F_1}$ / $Pre.$ / $Rec.$ (\%)$^\uparrow$} & $\mathbf{MAE}$ / $rMSE$ $^\downarrow$   & \multicolumn{1}{c|}{$\mathbf{F_1}$ / $Pre.$ / $Rec.$ (\%)$^\uparrow$} & $\mathbf{MAE}$ / $rMSE$ $^\downarrow$   \\ \whline
IIM-VGG16                & \multicolumn{1}{c|}{41.6 / 42.7 / 40.5}  & 181.3 / 412.3  & \multicolumn{1}{c|}{25.1 / 44.1 / 17.5}  & 242.8 / 676.1 & \multicolumn{1}{c|}{24.9 / 44.1 / 17.4}  & 176.5 / 914.8 & \multicolumn{1}{c|}{23.0 / 50.0 / 15.0}  & 245.9 / 666.1 \\ \hline
+Sunny Aug               & \multicolumn{1}{c|}{42.6 / 42.3 / 43.0}  & 176.3 / 389.49 & \multicolumn{1}{c|}{25.6 / 44.6 / 18.0}  & 240.2 / 643.2 & \multicolumn{1}{c|}{25.2 / 44.3 / 17.6}  & 177.3 / 914.6 & \multicolumn{1}{c|}{23.2 / 47.7 / 15.3}  & 238.8 / 630.9 \\ \hline
+Rainy Aug               & \multicolumn{1}{c|}{40.5 / 35.6 / 46.7}                    & 197.9 / 350.8               & \multicolumn{1}{c|}{26.7 / 30.7 / 23.6}                    &  274.7 / 662.8             & \multicolumn{1}{c|}{28.6 / 30.4 / 27.1}                    & 208.6 / 913.6              & \multicolumn{1}{c|}{27.4 / 40.0 / 20.9}                    &  219.3 / 589.5             \\ \hline
+Snowy Aug               & \multicolumn{1}{c|}{41.0 / 38.0 / 44.6}                    & 195.3 / 392.6               & \multicolumn{1}{c|}{26.1 / 35.3 / 20.7}                    & 254.9 / 660.5              & \multicolumn{1}{c|}{28.9 / 32.4 / 26.1}                    & 188.4 / 910.8              & \multicolumn{1}{c|}{26.9 / 44.3 / 19.3}                    & 224.4 / 614.1              \\ \hline
+Foggy Aug               & \multicolumn{1}{c|}{42.2 / 40.2 / 44.5}                    & 188.1 / 388.5              & \multicolumn{1}{c|}{26.7 / 37.7 / 20.6 }                    &    242.9 / 657.4           & \multicolumn{1}{c|}{28.2 / 34.1 / 24.0}                    & 179.7 / 910.2              & \multicolumn{1}{c|}{28.7 / 42.1 / 21.8}                    & 215.4 / 583.9              \\ \whline
\end{tabular}
}}
\end{table*}
\subsubsection{Setting\&Results}
Later, we conducted ablation analysis experiments through enhancement in the data scarcity scenario. 
As shown in Table~\ref{tab:Ablation}, each component of our model contributes significantly to its overall performance. Notably, the Isolated Box module results in an improvement of $3.5$ percentage points when used with naive RGB projection, and $1.6$ percentage points when combined with our novel Object Embedding technique, compared to the point representation approach. Moreover, our proposed IOE provides a substantial enhancement, boosting performance by $4.3$ percentage points over the baseline model. The inclusion of ISE further increases the model's efficacy, adding an additional improvement of $0.3$ percentage points. Lastly, the integration of the Position Shortcut module contributes an increase of $0.4$ percentage points. These results underscore the effectiveness of each proposed module in refining the model's performance. 
By combining all modules, our method achieved the best Crowd Localization Performance and the smallest Counting Error. 

{
\subsubsection{Further intuitive explanation}

The choice of IOE + ISE is driven by the failure of existing representations to simultaneously satisfy the scale and location control. Below, we elaborate on how each component fills critical gaps:

\paragraph{IOE: Fixing Spatial and Topological Limitations}
IOE addresses the \textit{spatial precision} and \textit{topological integrity} failures of points, density maps, and bounding boxes:
\begin{itemize}
    \item \textit{Shrunk Non-Overlapping Boxes}: Unlike standard bounding boxes, IOE uses iteratively shrunk boxes (adapted from IIM~\cite{IIM}) to eliminate overlaps in dense scenes. This ensures each instance occupies a unique spatial region, allowing the diffusion model’s cross-attention to map exactly one instance to one region (see Fig.~\ref{fig:Modality}: IOE avoids the ``merged heads'' of bounding boxes).
    \item \textit{Latent Alignment}: IOE uses PCA-reduced CLIP embeddings (Eq. 5) instead of raw RGB values. This aligns the control map’s latent space with Stable Diffusion’s VAE latent space, reducing the projection learning burden to a linear transformation (vs. non-linear for density maps). 
    \item \textit{Granularity Balance}: IOE avoids the ``too sparse'' issue of points (insufficient cues) and ``too blurred'' issue of density maps (ambiguous boundaries) by using region-wise embeddings. Each shrunk box provides enough spatial context to generate a complete instance without over-constraining appearance diversity (e.g., hair color, clothing).
\end{itemize}

\paragraph{ISE: Disentangling Scale from Location}
ISE solves the \textit{scale disentanglement} problem that plagues all prior representations:
\begin{itemize}
    \item \textit{Explicit Scale Encoding}: Unlike bounding boxes (scale = box size) or density maps (scale = density magnitude), ISE embeds scale as a separate cosine-sine vector. This breaks the ``location → scale'' bias (e.g., background = small) and allows users to define arbitrary scale-location pairs (e.g., ``large heads in the background''—impossible with prior methods).
    \item \textit{Smooth Generalization}: Cosine-sine embedding ensures a continuous scale space. For example, the model can interpolate between 10 pixels$^2$ (small) and 50 pixels$^2$ (large) without retraining, whereas linear encoding would require discrete thresholds (limiting flexibility).
    \item \textit{IOE Compatibility}: ISE is added directly to IOE’s latent vector, preserving spatial-topological information while injecting scale. This avoids the ``scale-spatial conflict'' of methods that modify spatial representations (e.g., resizing boxes) to encode scale— which corrupts location precision.
\end{itemize}
}

\subsection{Data Augmentation in Dataset Transfer Scenarios}\label{sec:Transfer}

\subsubsection{Experimental Setup}
We investigate dataset transfer across three scenarios: training on SHHA and testing on SHHB, QNRF, and NWPU, respectively. Comparative analysis is performed on crowd localization models, IIM and STEERER, assessing performance \emph{with} and \emph{without} the inclusion of generated images from DenseControl as augmentation.

\subsubsection{Results and Discussion}
As indicated in \tableautorefname~\ref{tab:Transfer}, our approach significantly improves the performance of IIM and STEERER algorithms in transfer scenarios. Specifically, the IIM algorithm's F1 score increases by approximately $5\%$, $4\%$, and $4\%$ across three scenarios, while the MAE is reduced by 10, 58, and 26, respectively. These enhancements are likely due to the diverse data generated by our method, underscoring its potential for robust crowd analysis across different datasets.

\subsection{Data Augmentation under Weather Transfer Scenario}\label{sec:weather}

\subsubsection{Setting}
We conduct an investigation into weather transfer across four distinct scenarios: training on sunny scenes and subsequently testing on sunny, rainy, snowy, and foggy conditions. Following this, we employ DenseControl to synthesize images based on instructions specific to these four weather scenarios, which serves as a method of data augmentation for the crowd localization model, IIM. The effectiveness of this approach is evaluated by assessing the model's performance with images generated from varying instructions.
\subsubsection{Reult Analysis}
From \tableautorefname~\ref{tab:Weather}, we have the following observations:
1) In most cases, data augmentation through our method leads to improvements in Crowd Localization Performance. 
Targeted enhancement for specific scenarios (e.g. +Rainy Aug for Sunny$\longrightarrow$ Rainy) can yield the best transfer effects, which aligns completely with our understanding.
2) In terms of head counting, our method may have a negative impact, as discussed earlier. 
This is primarily due to the noise present in the generated data.

\section{Conclusion and Future Work}

In this study, we introduce DenseControl, a pioneering approach for the controllable synthesis of dense crowd images at the instance level. DenseControl addresses key challenges in embedding control signals into the generation process, notably through the novel Isolated Object Embedding map and Implicit Scale Embedding. These innovations simplify spatial control and enhance scale precision respectively.
Our development of the Position Shortcut mechanism further refines the projection process, optimizing cross-attention within diffusion models and further improving synthesis quality. Rigorous testing across varied conditions confirms DenseControl superior performance in generating high-quality images and its effectiveness in practical applications like data scarcity augmentation and weather generalization.
Moving forward, we see immense potential in expanding DenseControl for instance-level attribute editing, leveraging our advancements in spatial and scale accuracy. 

{
\section*{Limitations}
\subsection{Dependence on the Quality of Instance Annotations}

DenseControl requires \textbf{accurate instance-level }annotations (location + approximate head scale) to construct the control map.
If the annotations contain large errors or missing labels, the synthesis quality may degrade.

\subsection{Dependence on the Basic Generation Capacity of Foundational Diffusion Model}

DenseControl is built on top of a Stable-Diffusion–based foundation model. Therefore, the overall image fidelity and diversity are partly constrained by the capability of the underlying diffusion backbone. This means that when the foundational model has limited expressiveness in certain scenes (e.g., rare artistic domains), our output quality may also plateau.

\bibliographystyle{IEEEtran}
\bibliography{ref}

\end{document}